\documentclass{vgtc}                          

\ifpdf
  \pdfoutput=1\relax                   
  \usepackage{graphicx}                
  \DeclareGraphicsExtensions{.pdf,.png,.jpg,.jpeg} 
\else
  \usepackage{graphicx}                
  \DeclareGraphicsExtensions{.eps}     
\fi%

\graphicspath{{figures/}{pictures/}{images/}{./}} 
\usepackage{subfig}
\usepackage{microtype}                 
\PassOptionsToPackage{warn}{textcomp}  
\usepackage{textcomp}                  
\usepackage{mathptmx}                  
\usepackage{times}                     
\usepackage{cite}                      
\usepackage{tabu}                      
\usepackage{booktabs}                  
\usepackage{multirow}
\usepackage{url}
\usepackage{balance}

\onlineid{1007}

\vgtccategory{Research}

\vgtcinsertpkg

\title{VIDES: Virtual Interior Design via Natural Language and Visual Guidance}

 \author{Minh-Hien Le\thanks{e-mail: 19120225@student.hcmus.edu.vn}\\ %
         \parbox{1.4in}{\scriptsize \centering University of Science \\ VNU-HCM, Vietnam} \\%
\and  
Chi-Bien Chu\thanks{e-mail: 19120002@student.hcmus.edu.vn}\\ %
      \parbox{1.4in}{\scriptsize \centering University of Science \\ VNU-HCM, Vietnam} \\%
\and 
Khanh-Duy Le\thanks{e-mail: lkduy@fit.hcmus.edu.vn}\\%
    \parbox{1.4in}{\scriptsize \centering University of Science \\ VNU-HCM, Vietnam} \\%
\and
Tam V. Nguyen\thanks{e-mail: tamnguyen@udayton.edu}\\%
    \parbox{1.4in}{\scriptsize \centering Dept. of Computer Science \\ University of Dayton, US} \\%
\and 
Minh-Triet Tran\thanks{e-mail: tmtriet@fit.hcmus.edu.vn}\\%
    \parbox{1.4in}{\scriptsize \centering University of Science \\ VNU-HCM, Vietnam} \\%
\and 
Trung-Nghia Le\thanks{e-mail: ltnghia@fit.hcmus.edu.vn (Corresponding author)}\\%
    \parbox{1.4in}{\scriptsize \centering University of Science \\ VNU-HCM, Vietnam} \\%
}

\teaser{
  \centering
  \includegraphics[width=0.8\linewidth]{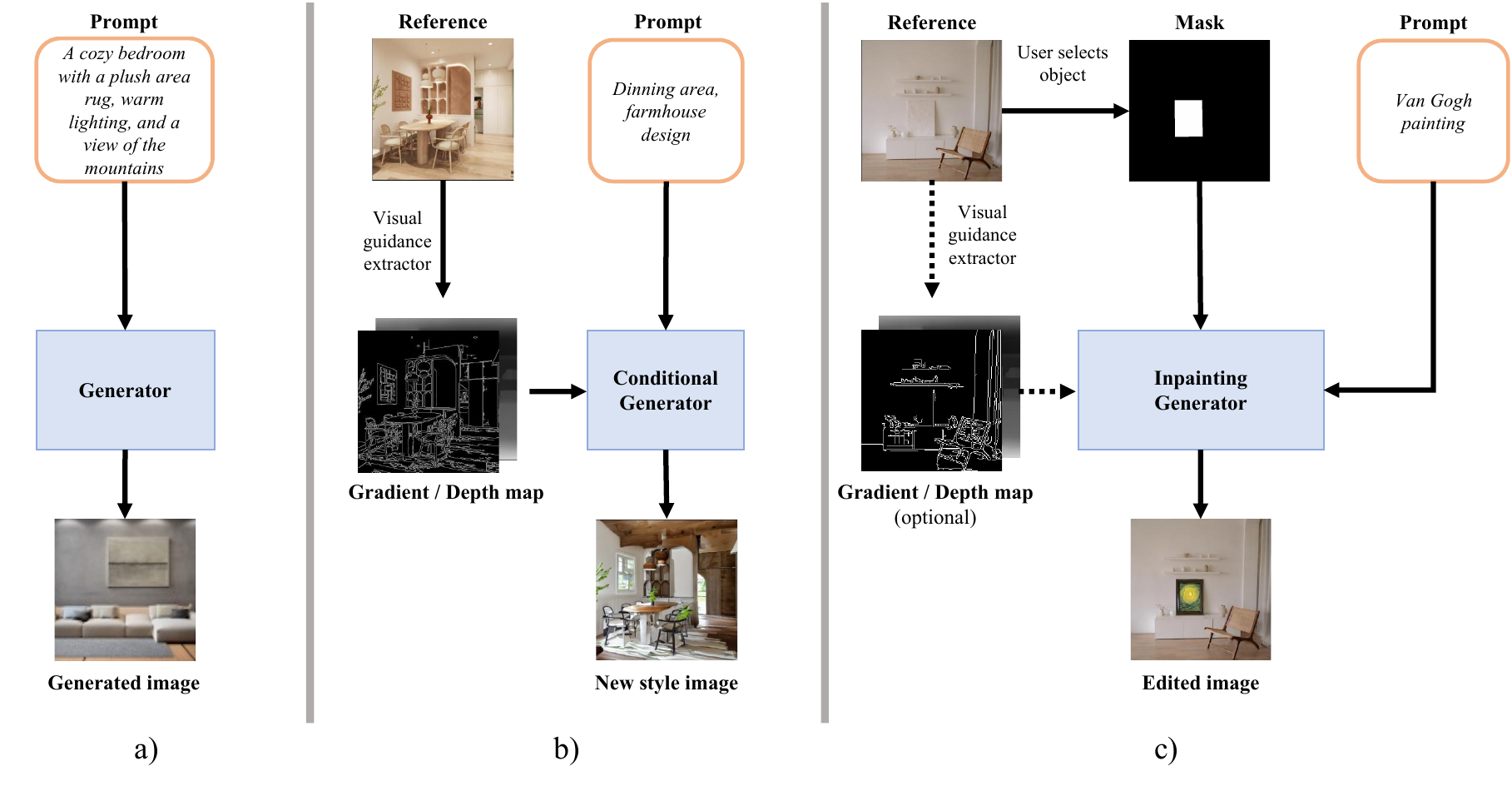}
  \caption{The overview of our proposed system. Its main components include (a) Indoor scene generation, (b) Style change, and (c) Object editing, including object removal and object replacement.}
  \label{fig:teaser}
}

\abstract{Interior design is crucial in creating aesthetically pleasing and functional indoor spaces. However, developing and editing interior design concepts requires significant time and expertise. We propose Virtual Interior DESign (VIDES) system in response to this challenge. Leveraging cutting-edge technology in generative AI, our system can assist users in generating and editing indoor scene concepts quickly, given user text description and visual guidance. Using both visual guidance and language as the conditional inputs significantly enhances the accuracy and coherence of the generated scenes, resulting in visually appealing designs. Through extensive experimentation, we demonstrate the effectiveness of VIDES in developing new indoor concepts, changing indoor styles, and replacing and removing interior objects. The system successfully captures the essence of users' descriptions while providing flexibility for customization. Consequently, this system can potentially reduce the entry barrier for indoor design, making it more accessible to users with limited technical skills and reducing the time required to create high-quality images. Individuals who have a background in design can now easily communicate their ideas visually and effectively present their design concepts.
}

\CCScatlist{
  \CCScatTwelve{Virtual reality}{Image generation}{Image editing}{Interior design}
}

\begin{document}

\maketitle


\section{Introduction} 

In recent years, there has been an increasing demand for interior design \cite{statista2017, technavio2022, yahoo2023}, particularly in high-end real estate properties such as luxury villas and penthouses. Moreover, design agencies are required to provide a variety of design concepts for the same room to the diverse preferences of their clients. These design concepts must fulfill aesthetic requirements and align with the client's vision. However, the current process of interior design relies heavily on manual labor and lacks the incorporation of AI capabilities, resulting in time-consuming and costly endeavors. Furthermore, this process necessitates the involvement of professionals with expertise in image editing to effectively visualize the design concepts. Consequently, image editing becomes a challenging task for individuals lacking the necessary design skills. Existing image editing software, such as Adobe Photoshop and AutoCAD, often prove arduous for users without a design background, as they demand significant time, effort, and proficiency to achieve desired outputs. Additionally, these traditional image editing applications do not provide the flexibility to incorporate user inputs, such as text description, object edges, or depth information, which can hinder the realization of the final design outputs.

Recent advancements in artificial intelligence, particularly in the field of image generation and image editing, have paved the way for rapid and straightforward interior design processes. These advancements primarily rely on two popular approaches: Generative Adversarial Networks (GANs) \cite{zhu2016, abdal2020image2stylegan, jahanian2020steerability} and Diffusion Models \cite{rombach2021, cheng2023diss, lugmayr2022repaint}.
GAN-based methods have demonstrated high-fidelity image synthesis and fast inference speed, although they often encounter challenges with training instability and mode collapse. On the other hand, Diffusion Models offer the advantage of producing high-quality and realistic images, surpassing the capabilities of GANs. Moreover, the integration of large language models (LLMs) into diffusion models facilitates the straightforward expression of user ideas. For instance, various popular models have gained significant attention on the internet such as  Imagen \cite{saharia2022}, DALL-E \cite{ramesh2021a}, Stable Diffusion \cite{rombach2021} can generate images based on user prompts or with the aid of reference images, enabling a more intuitive and user-friendly design process.

In this paper, we aim to develop an AI-based solution that allows standard users to modify interior images in a virtual reality (VR) environment without requiring technical skills. Leveraging cutting-edge diffusion technologies, our \textbf{V}irtual \textbf{I}nterior \textbf{DES}ign (\textbf{VIDES}) system empowers users to design interior concepts based on natural language instructions and visual guidance. The proposed system enables the ability to generate new design concepts and change interior style according to the user input guidance (see \autoref{fig:teaser}). It realizes new design concepts by intelligently interpreting user prompts and visual guidance. It also has the ability to change the room's style while preserving the original layout, ensuring a harmonious blend of user preferences and existing spatial arrangements.

Moreover, VIDES system also assists in modifying interior objects easily, such as object removal and object replacement (see \autoref{fig:teaser}). By enabling users to define their conceptual guidance (e.g., image gradient or depth map), our VIDES system offers more customization than traditional image editing software. This can be particularly useful for professionals (i.e., designers) and businesses (i.e., housing agencies) that quickly require customized images to suit their branding and marketing needs. Leveraging these visual guidances, our system streamlines replacing interior objects in images easily, eliminating the laborious process of copying, cutting, and pasting objects as in conventional image editing software. 


We further enhance users' interactions by enabling the selection and isolation of objects within images utilizing advanced segmentation models. Unlike the traditional approach of manually drawing detailed bounding lines that match the edges of the objects, which can be a time-consuming process taking hours, users can now choose objects by means of clicking or freely drawing bounding boxes around them. The selected objects can subsequently be edited while preserving the remainder of the images unaltered. Combining image segmentation and image generative models will significantly reduce the time and effort required to select and edit objects, providing a more efficient and user-friendly experience.

\begin{figure}[t!]
    \centering
    \includegraphics[width=\columnwidth]{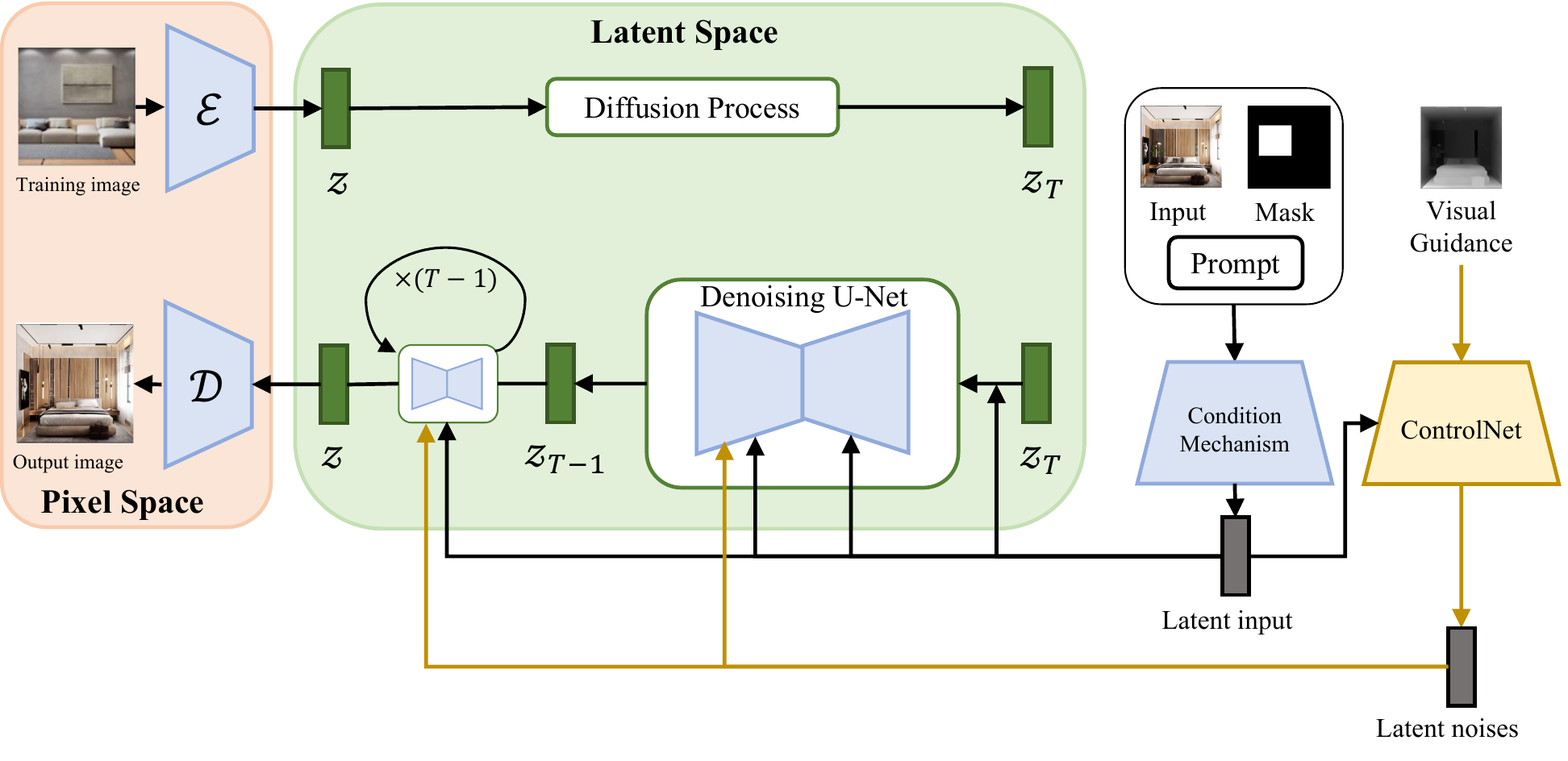}
    \caption{Overview of our generative AI framework. The network architecture consists of Stable Diffusion~\cite{rombach2021} inpainting with ControlNet~\cite{zhang2023adding} plugin.}
    \label{fig:ControlNetInpaining-arch}
\end{figure}

\begin{figure}[t!]
    \centering
    \includegraphics[width=0.8\columnwidth]{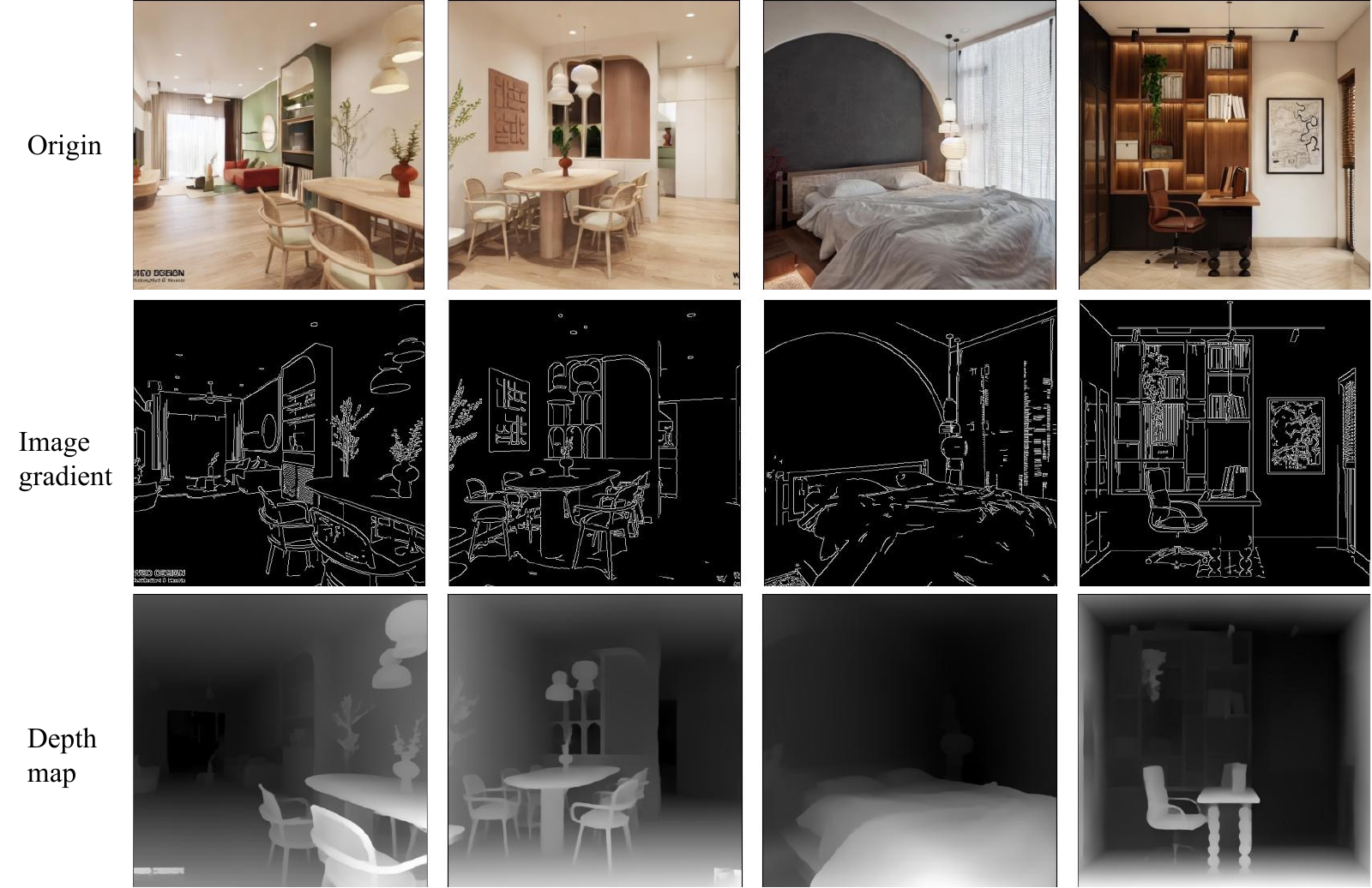}
    \caption{Examples of two conditions used for training our system.}
    \label{fig:condition-example}
\end{figure}

Beyond that, VIDES enables users to reconstruct the spatial representation of a room in a 3D format based on the provided 2D design image. This capability allows users to explore their design results from multiple perspectives, enhancing the overall user experience.

In order to achieve higher-quality results in the domain of interior images, we synthesized a comprehensive dataset of indoor scenes. Leveraging the capabilities large language model, we generated descriptive captions to generate new images via diffusion models. Although our system is fully trained on only our synthesized dataset, experimental results show that the generated indoor scenes exhibit a high level of realism and maintain a high-quality standard. The user study experiments have demonstrated that VIDES empowers users to focus on their creative vision without requiring specialized skills in image editing. Additionally, aside from the experimental results, VIDES has also contributed to inspiring users with new design ideas they had not previously considered. 

Our contributions lie in six-fold:
\begin{itemize}
    \item We propose a system for Virtual Interior Design via Natural Language and Visual Guidance. This integration empowers users with enhanced control over image creation and editing while maintaining accessibility across diverse user groups.
    \item Our system empowers users to generate interior design images from textual prompts. Subsequently, users can customize various aspects, such as design style, color palette, and more, while preserving the original layout of the interior.
    \item Our system enables users to select an object within an image and subsequently delete or modify it according to their preferences.
    \item We provide a user-friendly interface designed to accommodate a diverse range of user profiles, facilitating their interaction with the system within a virtual reality (VR) environment.
    \item We propose a novel synthesized dataset comprising high-quality interior images accompanied by corresponding textual descriptions. Although the system is fine-tuned solely on this dataset, it is capable of producing realistic results compared to real-world interior works.
    \item Through experimentation, real-world users have highly rated the system for its user-friendliness and the quality of its output.
\end{itemize}

\section{Related Works}
\subsection{Diffusion Models}

Diffusion models \cite{sohl-dickstein2015, croitoru2023, dhariwal2021, ho2020, nichol2021} are a class of generative probabilistic models that aim to approximate a data distribution. Recently, diffusion models have risen to prominence as state-of-the-art image generators due to their proficiency in learning complex distributions and diverse, high-quality images. By incorporating guiding input channels such as semantic layout and category label, diffusion models can be trained to achieve successful conditional image generation \cite{dhariwal2021, wang2022, sehwag2022}. Particularly, diffusion models that generate images from text prompts have been gaining significant popularity on the Internet recently and have demonstrated groundbreaking synthesis capabilities such as Imagen \cite{saharia2022}, DALL-E \cite{ramesh2021a}, Stable Diffusion \cite{rombach2021}.

Recent conditional diffusion models~\cite{avrahami2022, amit2022segdiff, meng2022sdedit, cheng2023diss, nichol2022glide, lugmayr2022repaint} have also shown superiority over GAN-based methods~\cite{zhu2016, abdal2020image2stylegan, jahanian2020steerability} in the image editing realm. They are utilized for reconstructing desired regions in the image based on various additional conditions such as sketch, stroke, composition, and realism. Particularly, Stable Diffusion and ControlNet \cite{rombach2021, zhang2023adding} has emerged as an approach to tackle the inpainting task by treating it as a general conditional image generation task and achieving significant results.

Due to diffusion models~\cite{rombach2021, zhang2023adding} capabilities to approximate complex data distributions and generate diverse, realistic, and high-quality images, we have chosen diffusion models as the primary building blocks of VIDES for synthesizing and editing images.

\subsection{Interior Design Generation Softwares}
Recent software products that utilize advanced generative AI models for interior design generation and editing have gained significant attention due to their impressive features, user-friendly interfaces.

RoomGPT \footnote{\url{https://www.roomgpt.io}} and ReRoom AI \footnote{\url{https://reroom.ai}} are among the first application that integrate generative AI and solely focuses on interior design. Their approaches allow users to harness the power of diffusion models to transform the style and theme of their uploaded room images. While allowing users to experiment with different design aesthetics and transform their living spaces, they remain relatively simple applications with a sole focus on changing room styles. The applications provide a limited list of predefined room themes, potentially limiting creative options. Hence, it falls short of meeting the needs of architects and interior designers who require more advanced features for their projects such as new room generation and object editing. While a noteworthy advancement in AI and interior design, RoomGPT and ReRoom AI are better suited for casual users seeking quick design changes rather than professionals in search of comprehensive, tailored solutions.

\begin{figure}[t!]
    \centering
    \includegraphics[width=0.7\columnwidth]{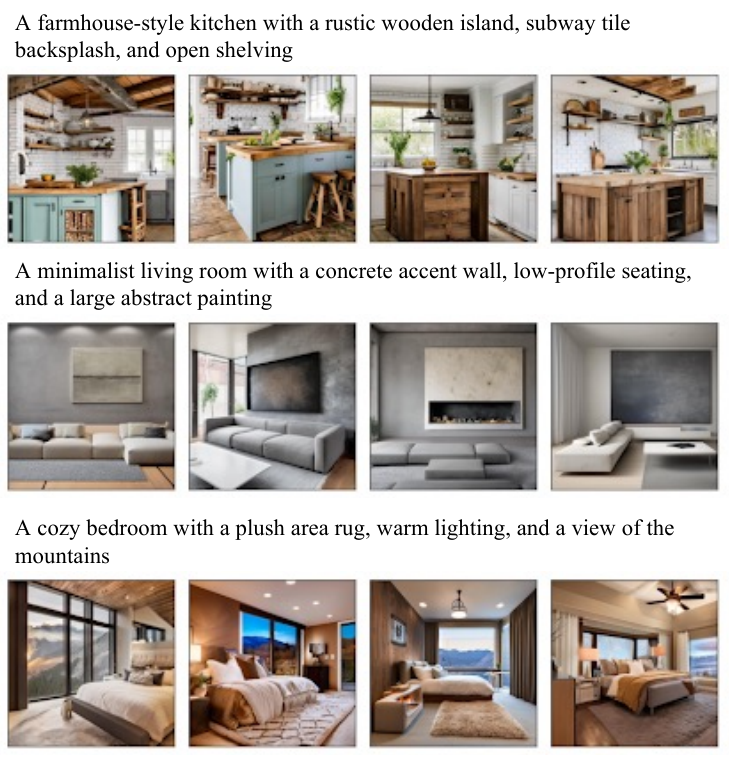}
    \caption{Examples of indoor scenes generated from prompts.} 
    \label{fig:dataset-example}
\end{figure}

\section{Proposed System}

\subsection{Generative AI for Interior Design}

\subsubsection{Overview}

We introduce a generative AI framework for image generation and editing, addressing the complexities of complex image manipulation tasks. Our approach leverages the strengths of multiple models to create a flexible and intuitive pipeline, empowering users to effortlessly generate and edit images. The framework consists of two main modules: an image generation module and an image editing module, which combine to enable rapid transformation and manipulation of images for various creative and practical applications. Both modules are based on the underlying architecture in \autoref{fig:ControlNetInpaining-arch}, differing only in the conditional input they are given.


\begin{figure}[t!]
    \centering
    \includegraphics[width=0.7\columnwidth]{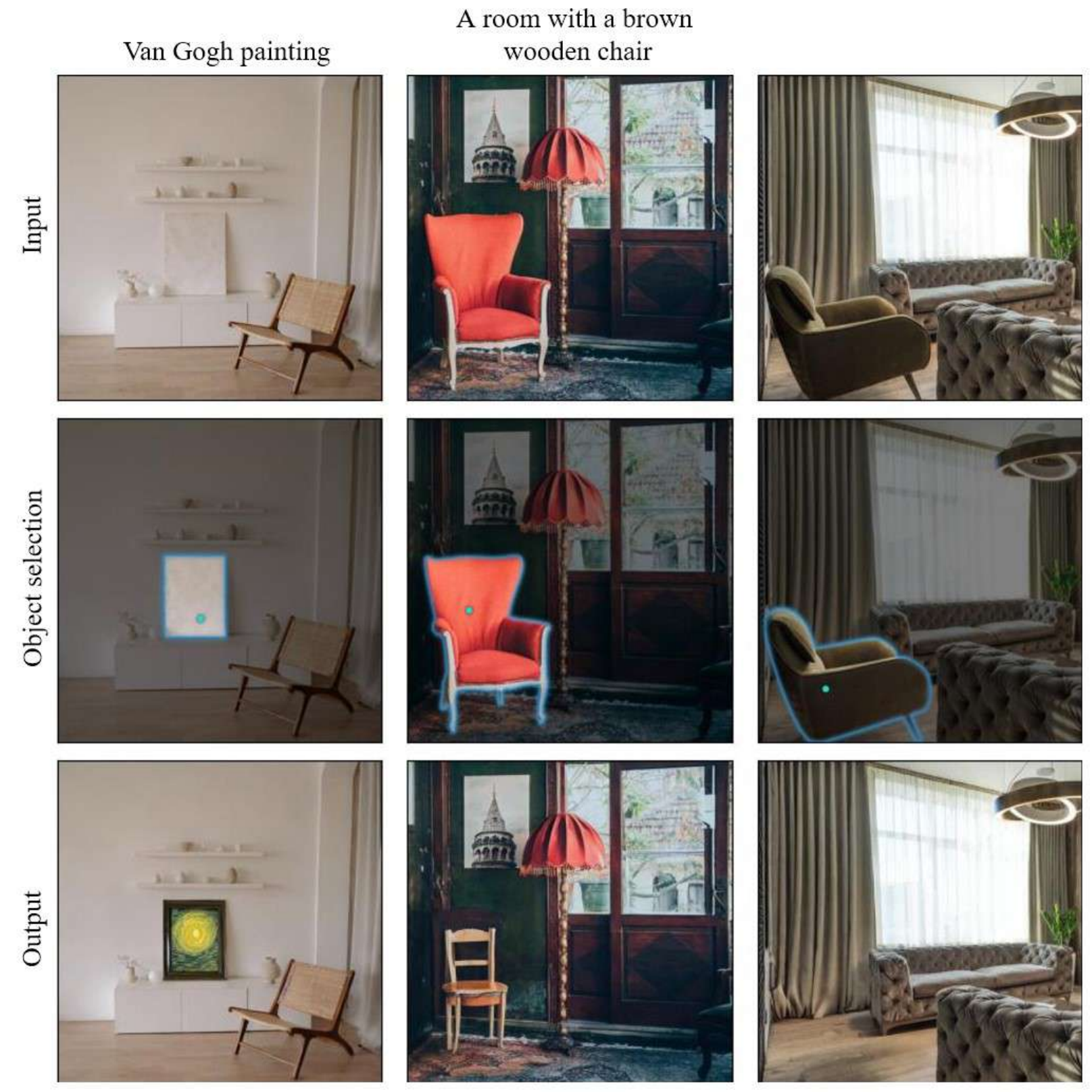}
    \caption{Examples of object replacements using prompts and visual guidance. The right column does not have a prompt, standing for object removal.}
    \label{fig:inference-object}
\end{figure}

\begin{figure}[tb]
    \centering
    \includegraphics[width=0.8\columnwidth]{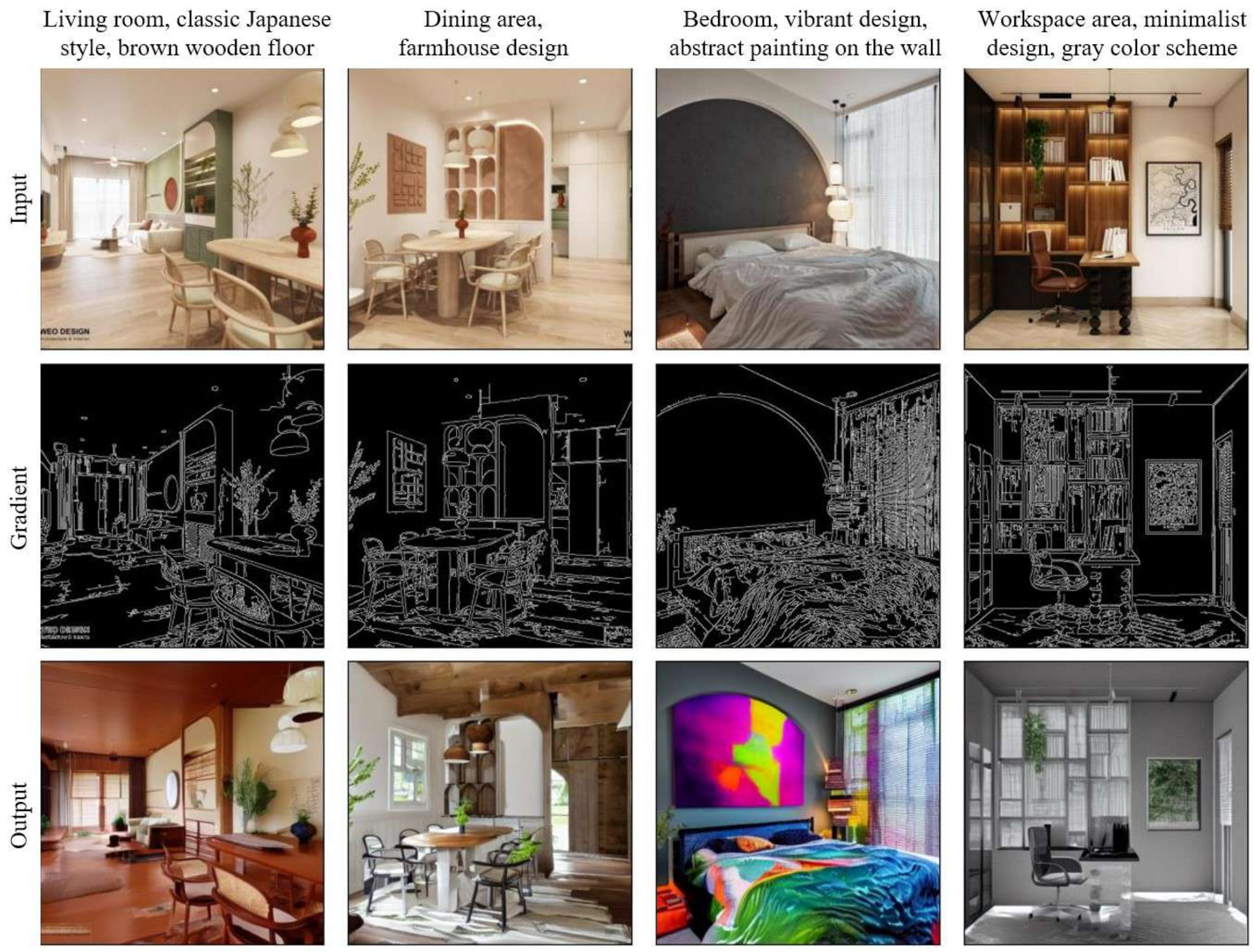}
    \caption{Examples of changing style using prompts and guided image gradient. The original layout is maintained.}
    \label{fig:inference-canny}
\end{figure}

\begin{figure}[tb]
    \centering
    \includegraphics[width=0.8\columnwidth]{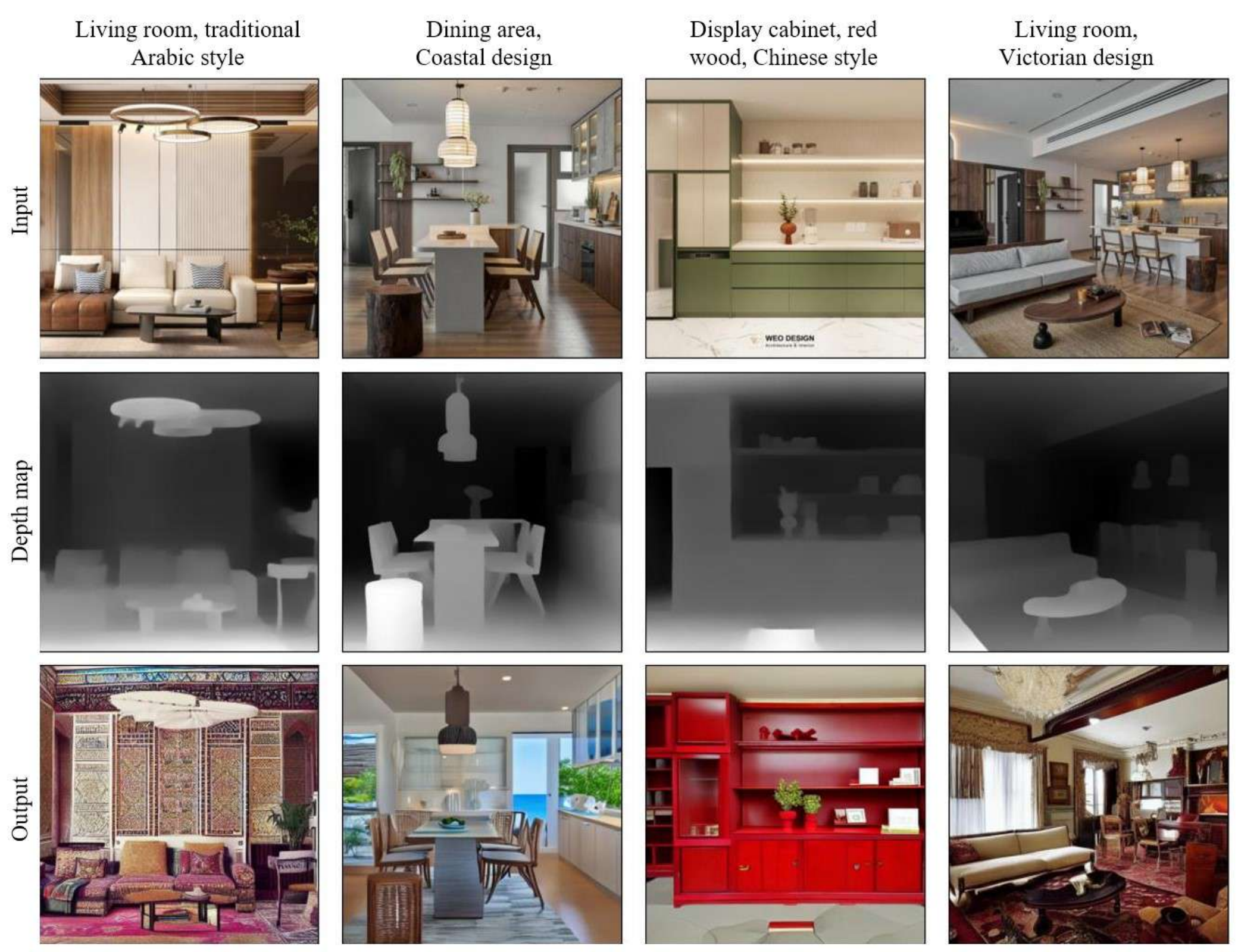}
    \caption{Examples of changing style using prompts and guided depth map. The original layout is maintained.}
    \label{fig:inference-depth}
\end{figure}

\subsubsection{Indoor Scene Generation}
\label{indoor-scene-generation}

\autoref{fig:ControlNetInpaining-arch} depicts the overview of our network architecture for indoor scene generation, which includes Stable Diffusion model \cite{rombach2021} integrated ControlNet \cite{zhang2023adding} plugin. ControlNet \cite{zhang2023adding} is a neural network structure that controls large image diffusion models, such as Stable Diffusion, by learning task-specific input conditions. It creates a trainable copy and a locked copy of the diffusion model's weights, enabling a new ability to accommodate various input visual conditions. Utilizing ControlNet, our framework provides users with a wide array of options to provide visual guidance, thereby boosting the generation of new images that align with their preferences and specific requirements. This robust framework allows users to control various aspects of the image generation process. It enables them to input and manipulate visual cues, such as initial images, text prompts, or a combination. By leveraging user-provided visual guidance, our system leverages advanced algorithms to synthesize highly customized image outputs, ensuring meeting user intent with the generated visual content. We remark that for image generation, the mask input is disregarded and is not utilized in the process. To demonstrate the capabilities of the indoor scene generation module, we provide \autoref{fig:teaser}, showcasing an example of generating a new indoor scene.


\subsubsection{Interior Object Editing}
\label{image-editing-component}



Besides producing complete images, Stable Diffusion \cite{rombach2021} also has the ability to edit specific areas within an image. In contrast to the procedure outlined in \autoref{image-editing-component}, the input includes an extra image for editing and a binary mask image that indicates the region requiring modification. These components will be processed using Condition Mechanisms, similar to how text prompts are used. 
Afterward, the image generation process occurs similarly to what was explained in \autoref{image-editing-component}. When users intend to preserve the object's form while making texture-only adjustments, the ControlNet \cite{zhang2023adding} is utilized to provide shape-related information like gradient or depth.

\subsection{Model Training}
\label{model-training}

We utilized the pre-trained Stable Diffusion model, which was already trained on large-scale datasets using the power of large language models. We only trained the ControlNet while freezing all other modules in the pipeline shown in \autoref{fig:ControlNetInpaining-arch}. 

We found that there are no available datasets for indoor scenes with both caption and high-quality images, which is necessary for training generative models. Therefore, we created a novel dataset for training our system. Firstly, we utilized ChatGPT to generate 1,500 sentence prompts describing indoor interior scenes. Subsequently, we employed diffusion models to generate corresponding images for each sentence prompt. Approximately 30 distinct images were generated for each prompt.

We fine-tuned two separate ControlNets with two types of conditions, such as image gradient and depth information (examples are shown in \autoref{fig:condition-example}). Particularly, we used Canny edge detection algorithm \cite{canny1986} to detect edges from the input image, resulting in image gradient, and MiDaS model \cite{Ranftl2022} to estimate the depth map from the input image.


\subsection{Visualization}


Examples of generated indoor scenes from prompts are presented in \autoref{fig:dataset-example}. The images generated exhibit a high level of realism and align well with the desired contexts (living room, kitchen, bedroom, etc.). Furthermore, the generated outputs closely adhere to the specified styles as described in the prompts (farmhouse, minimalism, cozy, etc.). The generated images from the same prompt also demonstrate diversity in terms of color palette, layout design, and perspective.

\autoref{fig:inference-canny} and \autoref{fig:inference-depth} illustrate some examples of changing design styles while maintaining the original layout. The outcomes demonstrate that image gradient and depth map provide valuable information about the layout, while the newly generated styles closely adhere to the requirements specified in the prompts.

\autoref{fig:inference-object} illustrates the results of editing a selected object in the scene. SAM model \cite{kirillov2023segment} is employed to generate a mask image highlighting the selected object. The proposed system demonstrates its capability to fulfill the requirements specified in the prompt. In addition, when the prompt is empty (last column), the model effectively removes the object from the image.







\subsection{Interactive Interface}

\begin{figure}[t!]
    \centering
    \includegraphics[width=0.8\columnwidth]{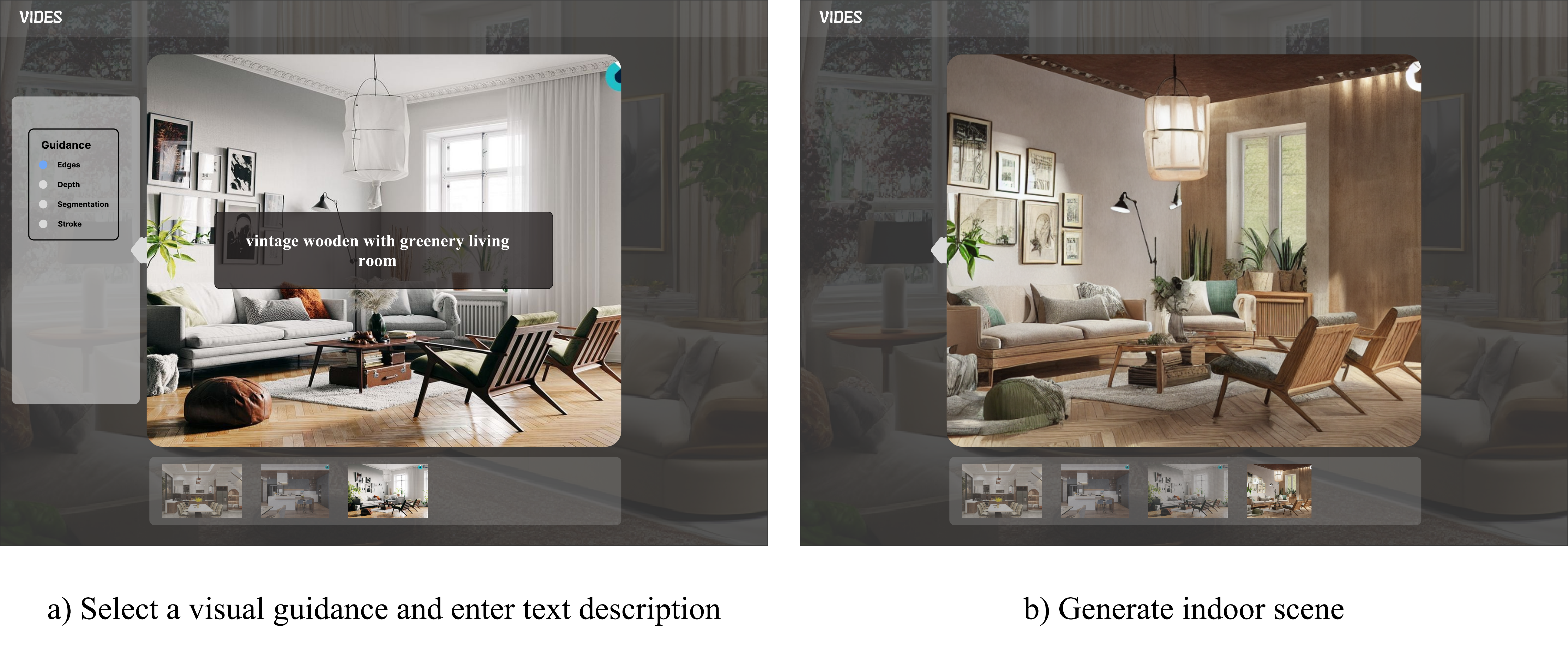}
    \caption{Interface of VIDES with components of indoor scene generation and interior style change. We note that the prompt block can be invisible to show the whole indoor scene.} 
    \label{fig:gui-generation}
\end{figure}
\begin{figure}[t!]
    \centering
    \includegraphics[width=0.8\columnwidth]{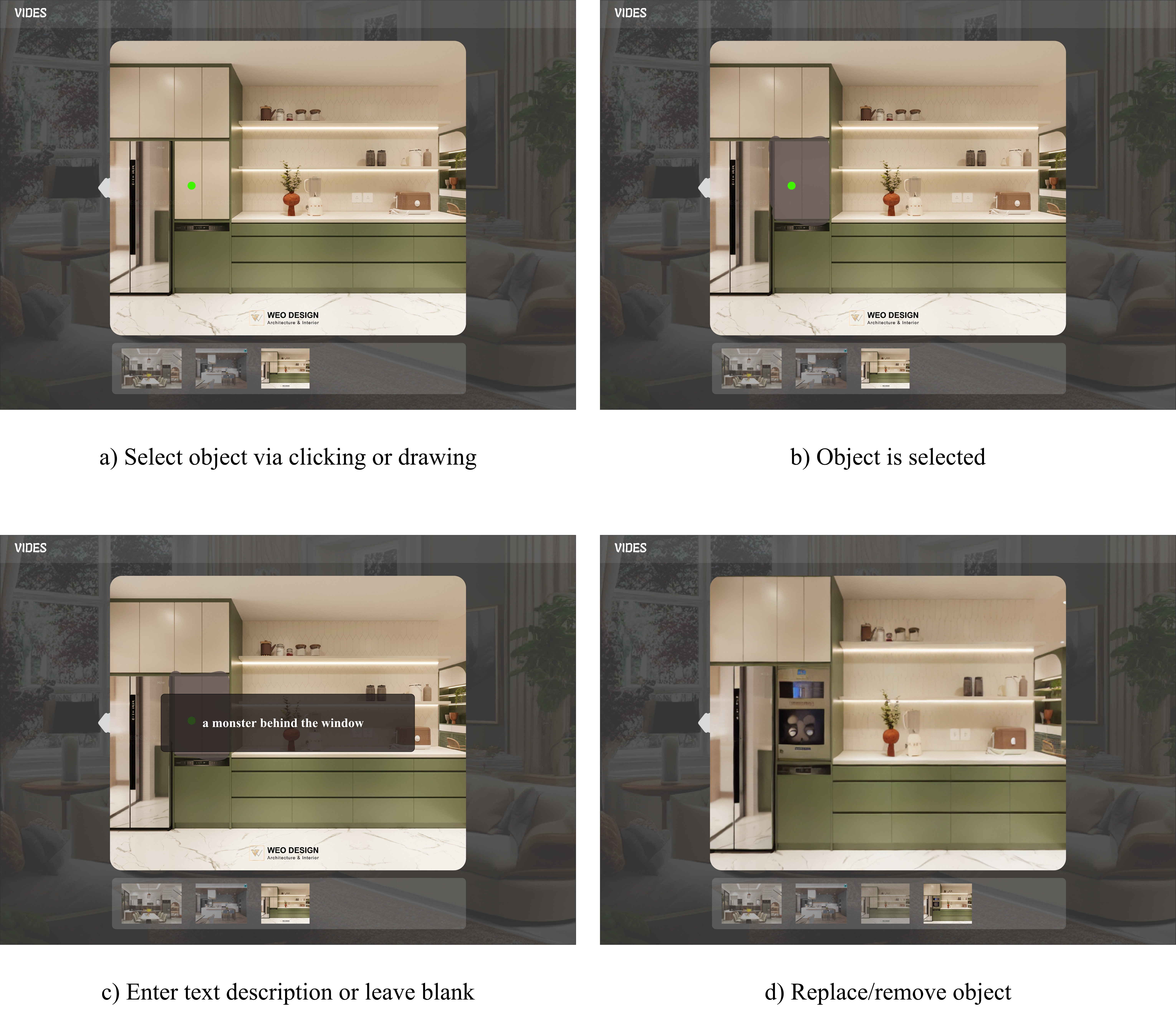}
    \caption{Interface of VIDES with components of object replacement and object removal. We note that the prompt block can be invisible to show the whole indoor scene.} 
    \label{fig:gui-edit}
\end{figure}

We develop a friendly interactive interface for interior design (see \autoref{fig:gui-generation} and \autoref{fig:gui-edit}). In addition, we also propose to leverage the strengths of multiple models, resulting in an interaction pipeline for generating and editing images. The interaction pipeline, illustrated in \autoref{fig:teaser}, is the foundation for empowering users to effortlessly generate and edit images, offering them a highly flexible and intuitive system. 



For indoor scene generation, users can provide either an initial image, a text prompt, or both, thereby guiding the generation process. The guidance is used to generate new images matching users' preferences. In addition, as shown in \autoref{fig:gui-generation}, users can upload an image of a room and provide a description of their desired interior style. The system then generates a new indoor style for that room but remain its layout.

Meanwhile, the interior object editing process is shown in \autoref{fig:gui-edit}. Users submit an initial image intended for editing, accompanied by a text prompt that serves as the replacement guidance. Additionally, users are expected to interactively select objects for editing by clicking on or drawing bounding boxes around them. The provided information is utilized to edit the chosen objects while keeping the remaining elements in the images untouched. We remark that if the guided prompt is not provided, the selected object is removed, i.e., image inpainting process.




\textbf{Object selection:} We utilize a recent computer vision foundational model, the Segment Anything (SAM) model \cite{kirillov2023segment}, to enable the ability to choose or isolate any desired object. Given user interactions (e.g., points, boxes, masks, text prompts, etc.), the model outputs the segmentation masks of the chosen objects. With the ability of SAM, we offer the ability to easily edit images interactively via clicking and drawing bounding boxes. 



\section{Evaluation}

\subsection{Datasets}

We utilized two image sources as ground truth. Particularly, we filtered out images in the MIT Indoor Scene dataset \cite{quattoni2009mit} and obtained a subset of over 5000 images belonging to classes related to indoor furniture; other unrelated classes were omitted. We also collected over 1400 real interior images of more than 30 high-end interior design projects from the Internet\footnote{\url{https://weo.vn}}.

\subsection{Image Generation}

To evaluate the effectiveness of our system in generating indoor scenes, we used the FID \cite{heusel2017fid} metric to measure the realism of the generated images compared with images in the real world. We generated 2000 interior images using our proposed system. The prompts were created by combining various combinations of room types (e.g., living room, bathroom, etc.), design styles (e.g., minimalist, farmhouse, etc.), and color schemes (e.g., monochromatic, warm, cool, etc.).

The comparative findings outlined in \autoref{tab:eval-fid} demonstrate the ability of our system to produce realistic indoor scenes. Notably, our generated results exhibit greater diversity than the collected interior images and closely resemble practical housing rooms. 



\begin{table}[t!]
    \centering
    \begin{tabular}{|l|c|c|} \hline
        & \multicolumn{1}{p{1.8cm}|}{\textbf{MIT Subset}} & \multicolumn{1}{p{1.8cm}|}{\textbf{Collected Interior Images}} \\ \hline
        MIT subset & - & 65.58 \\ 
        Collected interior images & 65.58 & - \\ 
        Our results & \textbf{51.51} & \textbf{65.33} \\ 
        \hline
    \end{tabular}
    \caption{Realism assessment of indoor images using FID metric \cite{heusel2017fid} (lower is better).}
    \label{tab:eval-fid}
\end{table}

\begin{table}[t!]
    \centering
    \begin{tabular}{|l|c|c|} \hline
        & \textbf{Image Gradient} & \textbf{Depth Map} \\ \hline
        MIT subset & 0.166 & 0.202 \\ 
        Collected interior images & 0.161 & 0.214 \\ 
        Our results & \textbf{0.161} & \textbf{0.180} \\ 
        \hline
    \end{tabular}
    \caption{Inpainting evaluation using different conditions on various indoor images. The performance is measured by LPIPS metric \cite{zhang2018lpips} (lower is better).}
    \label{tab:eval-lpips}
\end{table}

\begin{figure}[t!]
    \centering
    \includegraphics[width=0.9\columnwidth]{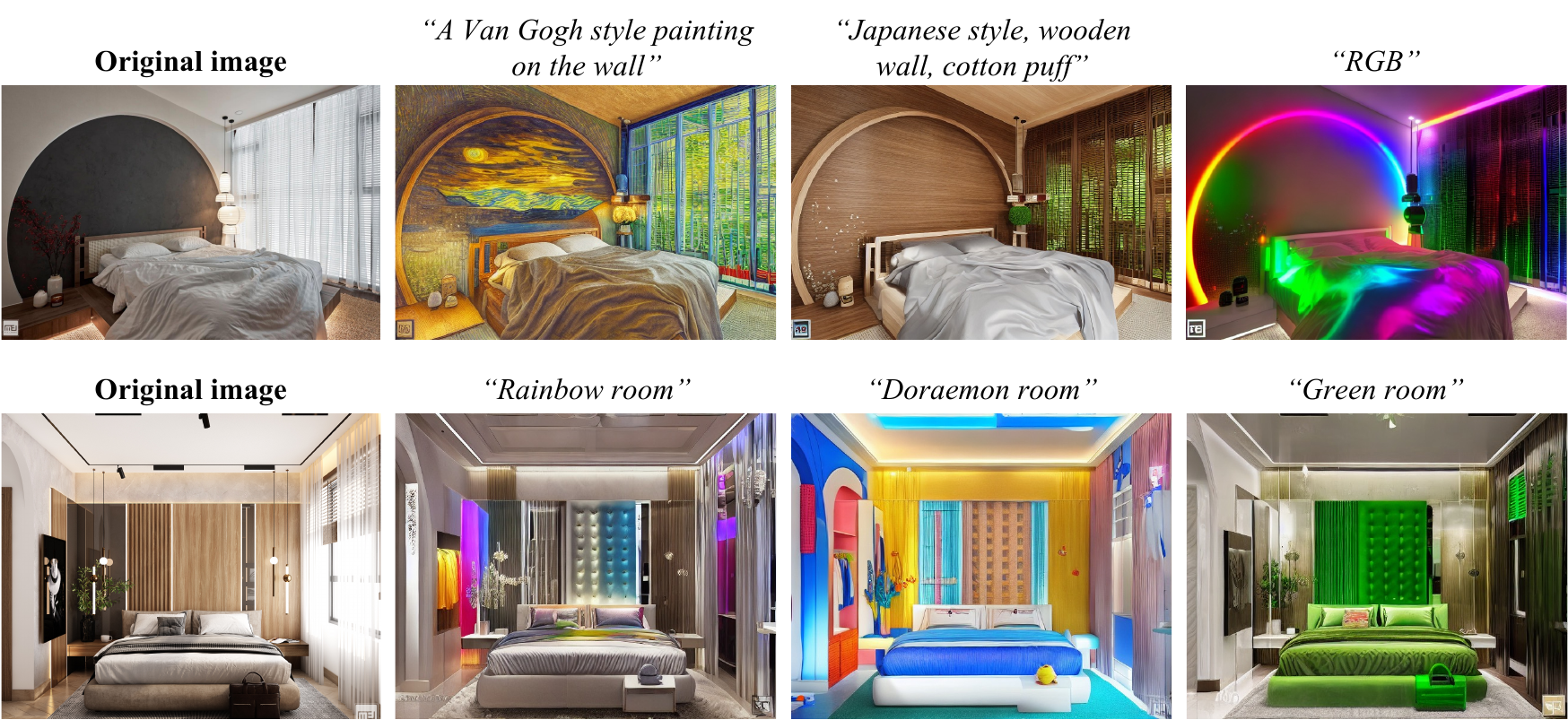}
    \caption{The results obtained when users performed indoor interior style change in the pilot study.}
    \label{fig:pilot-generate}
\end{figure}

\begin{figure}[t!]
    \centering
    \includegraphics[width=0.9\columnwidth]{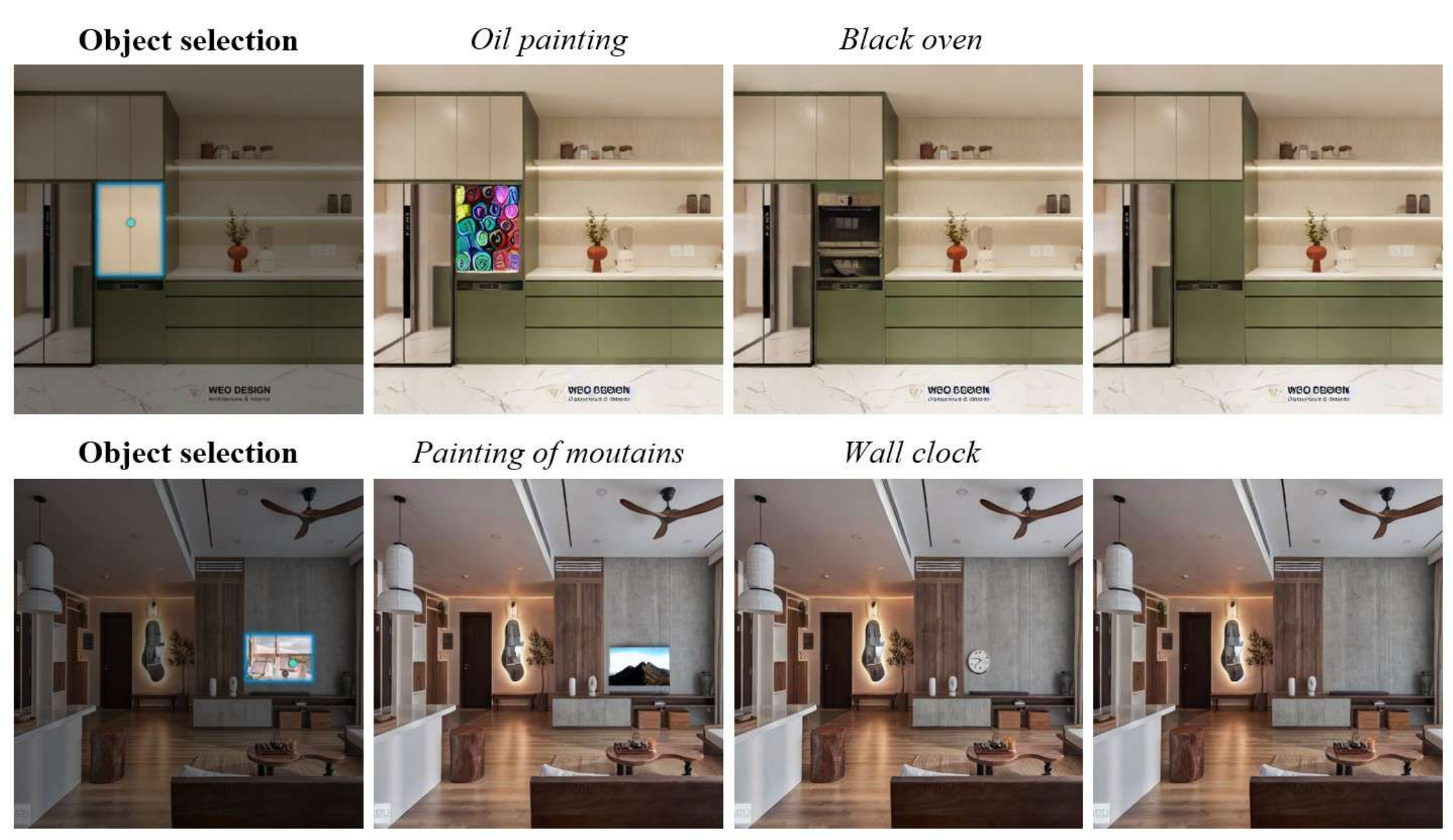}
    \caption{The results obtained when users performed object replacement and object removal (the last column) in the pilot study.}
    \label{fig:pilot-edit}
\end{figure}

\subsection{Object Editing}

To evaluate the effectiveness of our system in editing interior objects, we compared the input image before and after removing the selected object, using the LPIPS \cite{zhang2018lpips} metric. For each input image, we performed inpainting on a randomly selected region, covering 25\% to 64\% of the image area, separately using visual guidances such as image gradient and depth map. The experimental results in \autoref{tab:eval-lpips} indicate that image gradient is more helpful in reconstructing the scene than a depth map because it contains more visual information. The results also show that inpainting on our generated indoor scenes is much better than on other sources (i.e., MIT subset and collected interior images). 



\subsection{Pilot Study}




For our pilot study, we invited 12 participants including university students and researchers in the 18-44 age range who are familiar with design tools like Photoshop. Each participant took a 20-minute session, with each task allocated 10 minutes: room generation (generating a new room or changing the style of a room) and modifying objects within that room (removing objects or replacing them with new ones). Post-session, we conducted interviews on the system's effectiveness, receiving positive feedback and constructive improvement suggestions.

Participants had insightful discussions about our system. All reported high satisfaction with using VIDES. They were particularly impressed with the realistic and high-quality generated images, which captured about 83.3\% of their ideas, allowing them to explore and visualize different indoor styles and concepts. The users found the system helpful for those who needed help choosing a suitable interior styles for their rooms.

Users appreciated the system's convenience compared to Photoshop, finding it user-friendly even for individuals without a background in design, making it accessible to a wide range of users, especially indoor design agencies. Our intuitive interactive graphical user interface enhanced the system's usability, ensuring a seamless experience.

In feedback, participants provided constructive suggestions to improve VIDES. They noticed object replacement inaccuracies if given complex descriptions and suggested refining the algorithm. They also desired a comparison mode which allows them to quickly compare edited versions. This feature would enable users to track their design progress.

In response to this feedback, we plan to improve the object editing feature with complex descriptions. Additionally, we will add a comparison mode into the system, empowering users to evaluate their design choices and monitor their advancements. These enhancements elevate user experience and make VIDES more valuable for designers and enthusiasts. Some example results from real user experiences are shown in \autoref{fig:pilot-generate} and \autoref{fig:pilot-edit}.


\balance

\section{Conclusion}

We have introduced a generative AI-based system, namely VIDES, for advanced interior design to explore synthesized indoor scenes in a VR environment. Our system incorporates highly advanced pre-trained models for tasks such as image generation, image inpainting, and image segmentation. Additionally, to enhance the quality of results specifically in the domain of interior images, we fine-tuned the generative models on a novel dataset comprising synthesized interior images accompanied by descriptive captions. Experimental results show that our proposed system can represent a considerable contribution to the field, opening up new avenues for creative expression and enabling users to explore innovative possibilities in image editing and design. With VIDES, traditional limitations are shattered, allowing the standard users to embark on endless innovation.

Future exploration involves integrating 3D scene reconstruction techniques into our pipeline to enhance immersion and realism. Users can view 3D scenes of the modified room in a VR environment, opening up possibilities for architectural visualization and gaming. We also plan to expand more user visual guidance for image manipulation capabilities to provide greater control and personalization. Users can exert more influence over the image generation and editing process, resulting in outputs that better align with their creative vision.



\acknowledgments{
    This research is funded by University of Science, VNU-HCM, under grant number CNTT 2023-04.
}

\bibliographystyle{abbrv-doi}

\bibliography{ref}

\end{document}